\documentclass{article}

\usepackage{PRIMEarxiv}

\usepackage{hyperref}  
\usepackage{url}  

\usepackage[utf8]{inputenc} % allow utf-8 input
\usepackage[T1]{fontenc}    % use 8-bit T1 fonts
\usepackage{hyperref}       % hyperlinks
\usepackage{url}            % simple URL typesetting
\usepackage{booktabs}       % professional-quality tables
\usepackage{amsfonts}       % blackboard math symbols
\usepackage{nicefrac}       % compact symbols for 1/2, etc.
\usepackage{microtype}      % microtypography
\usepackage{lipsum}
\usepackage{fancyhdr}       % header
\usepackage{graphicx}       % graphics
\graphicspath{{media/}}     % organize your images and other figures under media/ folder

\usepackage{amsmath,amsfonts}
\usepackage{algorithmic}
\usepackage{algorithm}
\usepackage{array}
\usepackage{textcomp}
\usepackage{stfloats}
\usepackage{url}
\usepackage{verbatim}
\usepackage{cite}
\usepackage{amsmath}
\usepackage{booktabs}
\usepackage{graphicx}
\usepackage[caption=false,font=normalsize,labelfont=sf,textfont=sf]{subfig}
\usepackage{booktabs}
\usepackage{xcolor}
\usepackage{listings}
\usepackage{multirow}
\usepackage[most]{tcolorbox}
\usepackage[dvipsnames,svgnames,x11names]{xcolor}

\begin{document}

\title{SemanticAgent: A Semantics-Aware Framework for Text-to-SQL Data Synthesis}

\author{
  Qiang Gao$^{1\dagger}$,
  Zhenping Li$^{1\dagger}$,
  Anqi Zhuo$^{1,2}$,
  Yingxiao Zhao$^{1}$,
  Ruifang Zhao$^{1}$,
  Weibo Geng$^{1}$,
  Xiaosong Li$^{1}$\thanks{Corresponding author. 
  \texttt{lxiaosong2025@163.com}} \\[6pt]
  $^{1}$Center of Information Research, 
  Academy of Military Science, Beijing 100142, China \\
  $^{2}$School of Mathematics and Physics, 
  University of Science and Technology Beijing, 
  Beijing 100083, China \\[4pt]
  {\small $^\dagger$These authors contributed equally.}
}

\markboth{arXiv}%
{Gao \MakeLowercase{\textit{et al.}}: SemanticAgent for Text-to-SQL Data Synthesis}

% \IEEEpubid{0000--0000/00\$00.00~\copyright~2026 IEEE}

\maketitle

\footnotetext{Code available at: 
\href{https://github.com/lizhenping/SemanticSQL-Agent/tree/agent-publish}{https://github.com/lizhenping/SemanticSQL-Agent/tree/agent-publish}}

% \footnotetext[1]{$^\dagger$ These authors contributed equally to this work.}
% \footnotetext[2]{Correspondence to: Xiaosong Li 
% (\texttt{lxiaosong2025@163.com}).}

\begin{abstract}
Existing text‑to‑SQL synthesis pipelines still conflate executability with semantic validity: syntactic checks and execution‑based validation can retain queries that execute successfully while violating database semantics.
To address these limitations, we propose SemanticAgent, a semantic-aware synthesis framework. SemanticAgent organizes synthesis around three specialized modules: an analyzer, a synthesizer, and a verifier. Through a three-stage protocol of semantic analysis, stepwise synthesis, and diagnostic refinement. SemanticAgent transforms execution-based validation alone into a traceable reasoning process.
Our framework generates synthetic data that consistently outperforms prior synthesis methods under semantic-quality evaluation, leading to stronger downstream fine-tuning performance, especially on semantically demanding benchmarks.
\end{abstract}

\section{Introduction}\label{sec:introduction}

Text-to-SQL supports natural-language interaction with relational databases for non-expert users, and is increasingly used in practical scenarios such as business intelligence, healthcare analytics, and domain-specific question answering~\cite{shi2025survey}. 
However, developing effective text-to-SQL models requires large-scale, high-quality question–SQL training pairs. Constructing such pairs is costly, as each instance must be schema-grounded and carefully validated~\cite{zhang2023sciencebenchmark}. 
This challenge is particularly pronounced in domain-specific settings, where complex schemas and scarce domain expertise make it difficult to collect sufficient, annotation-ready data. For instance, clinical annotation required soliciting realistic questions from hospital staff, manually linking them to clinical schemas, and recruiting domain experts to verify each resulting SQL~\cite{lee2022ehrsql}. Data synthesis offers a promising alternative, reducing annotation costs and improving training coverage across diverse domains. 

Recent text‑to‑SQL synthesis pipelines leverage multiple guiding mechanisms to improve the quality of synthesized data~\cite{yu2021grappa,li2023resdsql,ye2023symgen,yang2024synthesizing,li2025omnisql,zhang2025exesql}. Specifically, seed examples provide reference patterns for candidate generation, and schema-level context ranging from raw DDL statements to enriched representations such as M-schema~\cite{liu2025} with descriptions and example values ground queries in the target database structure. Execution feedback then filters out non-executable SQL candidates. These mechanisms help maintain syntactic validity and structural alignment with the target schema. However, these signals are indirect indicators of semantic correctness and cannot ensure the semantic correctness of a synthesized question–SQL pair~\cite{sun2023sqlpalm,li2025sqlfactory}.

\begin{figure*}[!ht]
  \centering
  \includegraphics[width=.9\textwidth]{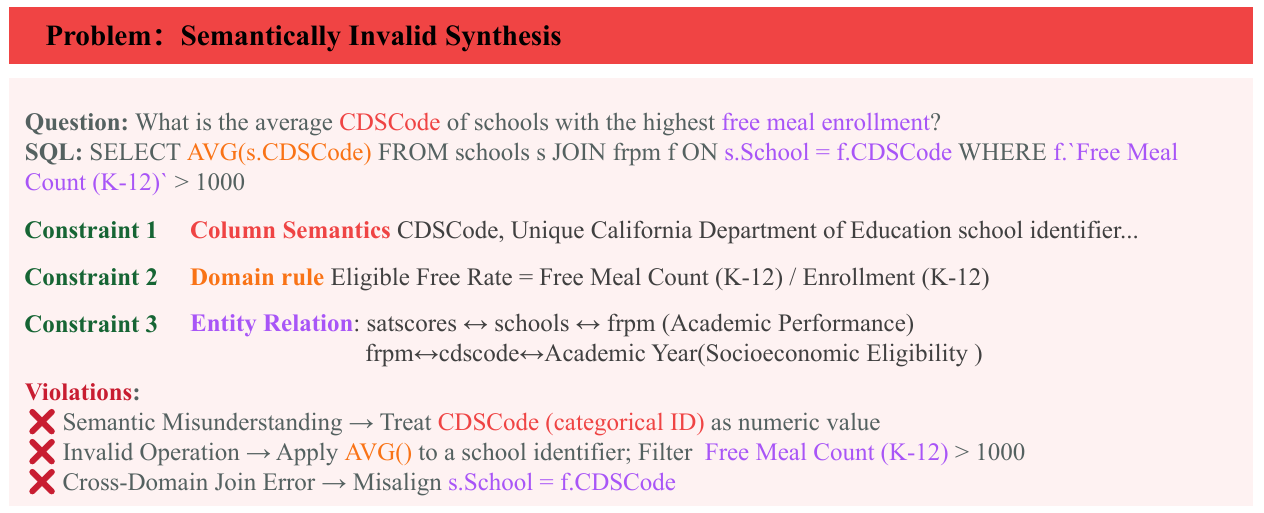}
  \caption{An example of a semantically invalid but executable SQL query. The aggregation AVG(CDSCode) over a school identifier violates domain semantics, yet passes execution-based validation.}
  \label{fig:semantic_error}
\end{figure*}

\begin{figure*}[ht]
  \centering
  \includegraphics[width=.7\textwidth]{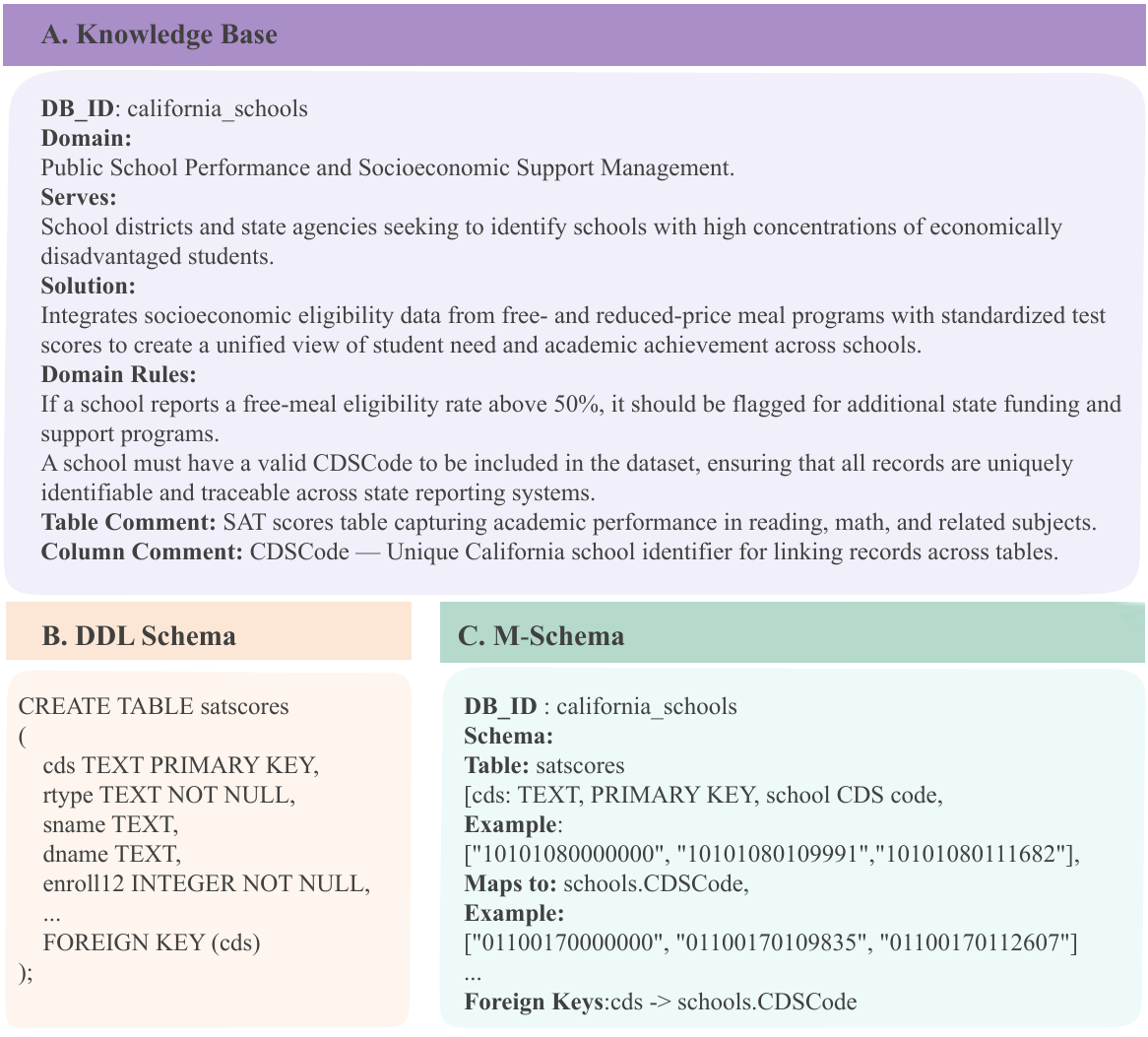}
    \caption{Comparison of database context used in LLM-based text-to-SQL. DDL (B) provides structural schema information, and M-schema (C) further enriches it with schema organization and example values, but both remain schema-level representations. In contrast, our knowledge base (A) makes domain-specific semantics and constraints explicit, enabling semantic checking during both generation and refinement.}
  \label{fig:knowledge}
\end{figure*}

The ability to analyze, generate, and verify relational queries is a cornerstone of high quality of data synthesis~\cite{shi2025survey,lee2022ehrsql}.
Realizing this capability requires addressing the inherent complexity of database semantics.
The semantic validity of a question–SQL pair is determined not by a single execution signal, but by the precise alignment of user intent, schema constraints, and domain semantics~\cite{liu2025}. A query may execute successfully, while its aggregation, join, or filter logic tells a semantically different story~\cite{shen2025study}. Consider \texttt{AVG(CDSCode)}: though the query executes without error, \texttt{CDSCode} denotes a school identifier, rendering the aggregation numerically valid but domain-meaningless(Fig.~\ref{fig:semantic_error}).
Accurately capturing such constraints throughout generation is the central challenge of high-quality data synthesis. While recent LLM-based methods have improved structural alignment and executability through demonstrations, enriched schema context(Fig.~\ref{fig:knowledge}), and execution feedback, a fundamental gap remains: these methods excel at syntactic constraint but often struggle with holistic semantic consistency throughout stepwise generation, leaving semantic deviations undetected and error sources untraceable~\cite{qu2025share}.

To address these challenges, we propose replacing implicit semantic modeling with explicit semantic supervision. We introduce SemanticAgent, a tool-augmented framework for text-to-SQL data synthesis. The framework redesigns the synthesis process through structured interaction among semantics-aware tools. SemanticAgent comprises three core tools: a Data Analysis Tool (DA), which acts as a domain analyst to identify business rules and semantic constraints governing the target schema; a Data Synthesis Tool (DS), which acts as a data author to compose questions, SQL queries, and operation rationales; and a Diagnosis Tool (DT), which functions as a semantic reviewer to inspect outputs and detect semantic inconsistencies. The overall process is organized into three structured phases: (1) Semantic Analysis, where the DA Tool derives domain-specific semantics and constraints from schema definitions to build a structured business knowledge base(Fig.~\ref{fig:knowledge} and App.~\ref{app:er-analysis}); (2) Controlled Authoring, where the DS Tool incrementally composes question–SQL pairs and rationales under explicit semantic supervision; and (3) Diagnosis, where the DT inspects outputs against domain-specific constraints, identifies error sources, and repairs semantically invalid samples.

Experimental results show that SemanticAgent produces higher-quality synthetic text-to-SQL data and leads to stronger downstream performance. For the downstream evaluation on BIRD under Qwen2.5-Coder-7B, models fine-tuned on SemanticAgent generated data outperform those fine-tuned on baseline synthetic data, with a 2.6 point improvement in execution accuracy (EX) over OmniSQL~\cite{li2025omnisql}. For the data quality evaluation, SemanticAgent-generated data also achieves higher semantic fidelity and diversity than data produced by baseline synthesis methods.

Our main contributions are summarized as follows:
\begin{itemize}
\item A Knowledge-Guided Framework for text‑to‑SQL Data Synthesis: We propose SemanticAgent, a tool-augmented agent framework that introduces explicit semantic supervision into text‑to‑SQL data synthesis, replacing reliance on implicit semantic assumptions.
\item Explicit Semantic Guidance throughout Synthesis: SemanticAgent organizes the synthesis pipeline into semantic knowledge extraction, instruction synthesis, and verification, incorporating domain knowledge and constraints to improve controllability and semantic consistency.
\item Empirical results show that SemanticAgent creates synthetic data with greater reliability and diversity, leading to stronger downstream text‑to‑SQL performance and more effective downstream learning.
\end{itemize}

\section{Related Work}\label{sec:related}

\subsection{Synthetic text-to-SQL Data Generation}
Manually annotating question–SQL pairs is costly and difficult to scale, making high-quality training data a persistent bottleneck for text-to-SQL systems. As a result, synthetic data generation has emerged as a scalable alternative, with recent work demonstrating its effectiveness for improving model performance across diverse domains~\cite{yang2024synthesizing,li2025omnisql}. Existing approaches improve synthetic text-to-SQL data along two dimensions: candidate generation and quality refinement.

On the SQL side, early approaches synthesize queries by instantiating predefined templates across schemas~\cite{yu2021grappa}, while later methods decompose queries into abstract syntax trees and recombine reusable components to produce structurally diverse candidates~\cite{wu2021,baumgartner2024synql,guo2025sqlforge}. On the natural-language side, early work relies on lexical substitution~\cite{yang2021hierarchical} or seq-to-seq question generation from SQL~\cite{guo2018question}. More recent approaches leverage LLMs with schema-aware prompting and in-context learning to produce richer and more varied questions~\cite{ye2023symgen,yang2024synthesizing,liu2024reformer,caferoglu2025singsql}.

Beyond generation, recent pipelines add a refinement stage to improve data quality. Ye et al.\cite{ye2023symgen} apply execution-based filtering to remove invalid candidates after generation. Yang et al.\cite{yang2024synthesizing} combine generation with critic models and preference supervision to correct errors from weaker generators. Liu et al.\cite{liu2024reformer} alternate generation and validation in an iterative loop to progressively refine data quality. Caferoglu et al.\cite{caferoglu2025singsql} assess question–SQL consistency by comparing questions against natural-language SQL explanations.
However, broader coverage and stronger refinement do not guarantee semantic correctness. Most existing pipelines focus on structural diversity or execution validity, leaving domain-specific semantics and constraints largely unaddressed.

\subsection{Semantic Inconsistency in Synthetic Text-to-SQL Data}

Despite their practical effectiveness, existing synthesis pipelines for text-to-SQL face a fundamental challenge: semantic inconsistency in synthetic question–SQL pairs. In many cases, a synthesized pair may remain executable and pass surface-level quality checks, yet still violate the semantic alignment between the question and the underlying database constraints. This challenge is particularly acute in domain-specific settings, where semantic validity depends not only on schema structure but also on domain rules and data constraints that surface-level verification cannot capture.
Recent synthesis pipelines partially address this problem through execution-based filtering or model-guided refinement~\cite{ye2023symgen,yang2024synthesizing,liu2024reformer,caferoglu2025singsql}. However, these approaches rely on indirect proxies—such as executability or model-based plausibility—rather than directly verifying semantic consistency against schema constraints and data values. Consequently, executable but semantically invalid pairs may survive filtering and introduce noise into synthetic training data.
This limitation suggests that improving synthetic data quality requires more than stronger generation or refinement—it demands explicit mechanisms for semantic grounding and verification. We therefore next review text-to-SQL techniques closely related to these directions, including methods for stronger schema grounding and methods that expose intermediate reasoning processes.

\subsection{Text-to-SQL Techniques Relevant to Explicit Semantic Verification}

Beyond synthesis, several lines of text-to-SQL research are relevant to explicit semantic verification, including work on external knowledge grounding and intermediate reasoning decomposition.
One line of work improves grounding through external knowledge and schema-aware representations. Prior work shows that external knowledge helps models capture domain-specific constraints beyond parametric knowledge~\cite{hong2024knowledge,han2024bridging}, while schema linking and retrieval methods improve alignment between natural-language mentions and schema elements~\cite{wang2025linkalign,glass2025extractive,fan2024confidence,guo2025multipattern,zhang2025structure}. These methods strengthen SQL grounding but are not designed to verify whether synthesized data is semantically consistent with schema constraints.
Another line of work introduces intermediate reasoning and decomposition into the generation process. Recent methods apply chain-of-thought prompting~\cite{wei2022chain,chang2023prompt}, reasoning decomposition~\cite{sui2025chain,dai2025parsql,jiang2026magma}, and clause-level parsing~\cite{xu2024cop,pourreza2025chasesql} to expose intermediate decisions, improving generation accuracy and interpretability. Although these approaches enable finer-grained analysis and more targeted correction, they focus on improving final SQL quality rather than verifying semantic consistency during data synthesis.
Overall, existing work advances grounding and reasoning for text-to-SQL generation, but explicit semantic verification during data synthesis remains largely unaddressed—particularly for domain-specific settings where schema constraints and data semantics are critical.

\section{Methods}
\label{sec:methods}

\begin{figure*}[!t]
 \centering
 \includegraphics[width=0.9\textwidth]{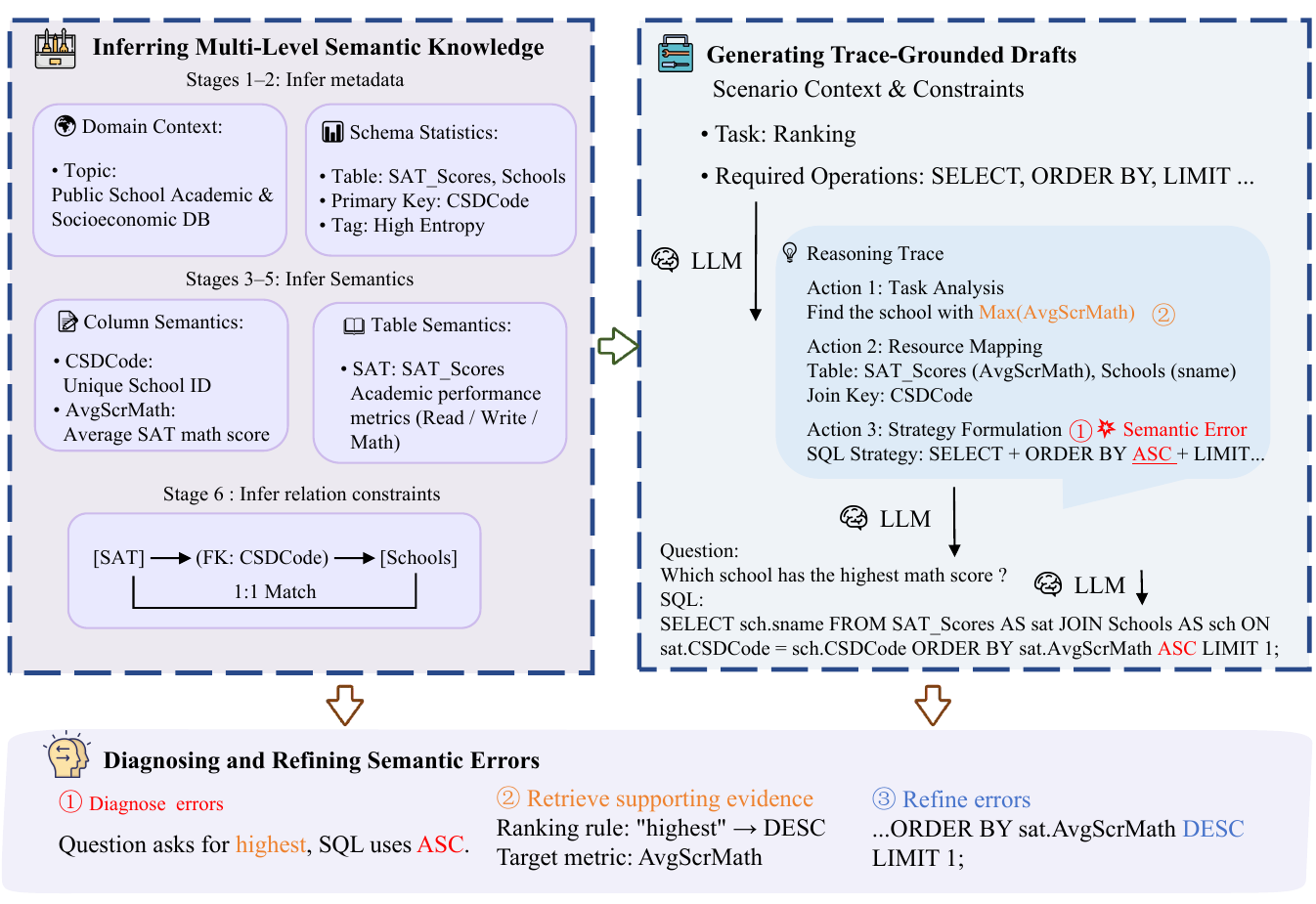}
\caption{Overview of SemanticAgent. Starting from database schemas, sampled table values, and synthesis controls, SemanticAgent (1) infers a multi-level semantic knowledge base, (2) generates trace-grounded question-SQL drafts under semantic guidance, and (3) diagnoses and refines semantic errors by retrieving supporting evidence from the knowledge base. The output is a semantically validated triple admitted to the synthetic corpus.}
 \label{fig:overview}
\end{figure*}

We aim to generate text-to-SQL training data that is semantically consistent with domain knowledge, rather than merely syntactically correct. To tackle this challenge, we develop \textbf{SemanticAgent}, a knowledge-guided synthesis framework that leverages a structured semantic knowledge base $\mathcal{K}$. The framework operates in three stages. First, it infers multi-level semantic knowledge from database schemas and instances. Second, it generates trace-grounded question--SQL drafts under semantic guidance. Finally, it diagnoses and refines semantic errors through execution and constraint-aware verification. Figure~\ref{fig:overview} illustrates the overall architecture.

\subsection{Problem Formulation}
\label{subsec:problem-def}

Given a relational database $\mathcal{D} = (\mathcal{S}, \mathcal{I})$ with schema $\mathcal{S}$ and instances $\mathcal{I}$, we aim to synthesize a training dataset $\mathcal{D}_{\text{syn}} = \{(q_i, s_i, r_i)\}_{i=1}^{N}$, where each sample consists of a natural language question $q_i$, an SQL query $s_i$, and a rationale $r_i$. A triple $(q, s, r)$ is semantically consistent if the question is faithfully aligned with the SQL query, the SQL query is syntactically valid and executable, and both the SQL query and the rationale conform to domain-specific constraints. Since schemas alone do not explicitly encode such semantic information, we construct a semantic knowledge base $\mathcal{K}$ from $\mathcal{I}$ to support both generation and semantic verification. See Appendix~\ref{app:training_format} for the detailed data format.

\subsection{System Architecture}
\label{subsec:architecture}

SemanticAgent consists of three components to support semantically faithful synthesis (Figure~\ref{fig:overview}). First, we extract domain knowledge from database instances $\mathcal{I}$ and organize it into a structured knowledge base $\mathcal{K}$, which captures entity semantics, type constraints, and entity relations (§\ref{subsec:semantic-construction}). Next, guided by $\mathcal{K}$, we generate triples $(q, s, r)$, where $q$, $s$, and $r$ denote the question, SQL, and rationale, respectively (§\ref{subsec:generation}). The rationale records the intermediate reasoning process, including semantically grounded table selection, type-constrained column operations, and SQL strategies aligned with domain rules. Finally, we validate the generated triples through execution and semantic verification (§\ref{subsec:refinement}). The validator uses $\mathcal{K}$ to detect violated constraints and trace the reasoning chain $r$, enabling targeted correction while preserving valid reasoning steps.

\subsection{Inferring Multi-Level Semantic Knowledge}
\label{subsec:semantic-construction}

Database schemas alone are insufficient to capture domain knowledge. To address this limitation, we extract a structured semantic knowledge base from the database through a six-stage hierarchical process, summarized in Table~\ref{tab:extraction-pipeline}.

The extraction process progresses from basic structural information to higher-level domain constraints and cross-table relations. The early stages extract schema metadata and domain-level information. The middle stages focus on intra-table semantics, including data type constraints, column-level properties, and table-level constraints. The final stage captures cross-table relations and dependencies. Each stage builds on knowledge extracted in earlier stages, enabling incremental semantic refinement.

We formalize knowledge extraction as a six-stage process. Let $\mathcal{K}_t$ denote the knowledge extracted at stage $t$. At each stage, an LLM-based extraction function takes stage-specific input context $\mathcal{X}_t$ and a prompt $\mathcal{P}_t$, and produces $\mathcal{K}_t$:
\begin{equation}
\mathcal{K}_t = f_{\text{LLM}}(\mathcal{X}_t, \mathcal{P}_t), \quad t \in \{1,\ldots,6\}
\end{equation}
where $\mathcal{X}_t$ may include the schema $\mathcal{S}$, a sampled subset of instances $\mathcal{I}_{\text{sample}} \subset \mathcal{I}$, and previously extracted knowledge $\mathcal{K}_{<t}$. The complete knowledge base is then given by $\mathcal{K} = \mathcal{K}_{1:6}$.

The extracted knowledge base $\mathcal{K}$ is organized for on-demand retrieval. The generation module queries $\mathcal{K}$ for domain guidance (§\ref{subsec:generation}), while the validation module retrieves constraint evidence for semantic verification (§\ref{subsec:refinement}). This design allows both modules to access relevant domain knowledge efficiently during synthesis.

\begin{table}[!ht]
\centering
\caption{Overview of the Six-Stage Semantic Extraction Process}
\label{tab:extraction-pipeline}

\begin{tabular}{lll}
\toprule
\textbf{Stage ($t$)} & \textbf{Input Context} & \textbf{Output ($\mathcal{K}_t$)} \\
\midrule
1. Schema Extraction & $\mathcal{S}, \mathcal{I}_{\text{sample}}$ & $\mathcal{K}_1$ (Metadata) \\
2. Domain Analysis & $\mathcal{S}, \mathcal{K}_1$ & $\mathcal{K}_2$ (Domain Constraints) \\
3. Field Type Analysis & $\mathcal{K}_{1}, \mathcal{I}_{\text{sample}}$ & $\mathcal{K}_3$ (Field Types) \\
4. Column Analysis & $\mathcal{K}_{1:3}$ & $\mathcal{K}_4$ (Column Semantics) \\
5. Table Analysis & $\mathcal{K}_{1:4}$ & $\mathcal{K}_5$ (Table Constraints) \\
6. Relation Analysis & $\mathcal{K}_{3}, \mathcal{K}_{4}, \mathcal{K}_{5}$ & $\mathcal{K}_6$ (Cross-Table Relations) \\
\bottomrule
\end{tabular}%

\end{table}

\subsection{Generating Trace-Grounded Drafts}
\label{subsec:generation}

Given the semantic knowledge base $\mathcal{K}$, we synthesize triples $(q, s, r)$ using structured prompts. The generation process is controlled along three complementary dimensions: domain context, analytical task type, and SQL complexity level (Table~\ref{tab:generation-config}). Conditioned on $(\mathcal{K}, \mathcal{C}, \mathcal{T}, \mathcal{L})$, the generator produces a natural language question $q$, an SQL query $s$, and a rationale $r$ that explicitly records the intermediate reasoning process.

\begin{table}[!ht]
\centering
\caption{Generation control parameters.}
\label{tab:generation-config}
\footnotesize
\setlength{\tabcolsep}{8pt}
\begin{tabular}{ll}
\toprule
\textbf{Parameter} & \textbf{Description} \\
\midrule
$\mathcal{C}$ & Domain context (e.g., sales, education) \\
$\mathcal{T}$ & Analytical task type (e.g., trend analysis, ranking) \\
$\mathcal{L}$ & SQL complexity level ($L_1$--$L_4$; Table~\ref{tab:complexity-levels}) \\
\bottomrule
\end{tabular}
\end{table}

We categorize queries into four structural complexity levels $\mathcal{L}$ (Table~\ref{tab:complexity-levels}), ranging from single-table queries ($L_1$) to modular compositions with CTEs ($L_4$). This design enables controllable synthesis over diverse SQL structures and supports balanced generation across different complexity levels.

\begin{table}[!ht]
\centering
\caption{SQL Complexity Levels}
\label{tab:complexity-levels}
\footnotesize
\begin{tabular}{ll}
\toprule
\textbf{Level} & \textbf{Query Features} \\
\midrule
$L_1$ & Single-table queries with filtering and aggregation \\
$L_2$ & Multi-table joins \\
$L_3$ & Nested subqueries and window functions \\
$L_4$ & CTEs and modular composition \\
\bottomrule
\end{tabular}
\end{table}

To formalize the generation process, we model each synthesis as governed by a generation scenario $\sigma = (\mathcal{C}, \mathcal{T})$, where $\mathcal{C}$ specifies the domain context (e.g., sales, education) and $\mathcal{T}$ defines the analytical task type (e.g., trend analysis, ranking). For a scenario $\sigma$ and a complexity level $\ell \in \mathcal{L}$, the prompt incorporates relevant constraints from $\mathcal{K}$ to generate:
\begin{equation}
(q, r) = f_{\text{que}}(\mathcal{K}, \sigma, \ell)
\end{equation}
where $q$ is the natural language question and $r$ is the reasoning trace. The constraints from $\mathcal{K}$ serve two purposes: (1) specifying valid field operations (e.g., aggregation-compatible columns), and (2) enforcing domain rules (e.g., excluding refunded transactions from revenue).

Given $(q, r)$ and the generation parameters $(\sigma, \ell)$, we then synthesize the SQL query as:
\begin{equation}
s = f_{\text{sql}}(q, r, \sigma, \ell, \mathcal{K})
\end{equation}

\subsection{Diagnosing and Refining Semantic Errors}
\label{subsec:refinement}

Given a generated triple $(q, s, r)$, we validate its executability and semantic faithfulness through an iterative feedback loop. Let $(q^{(k)}, s^{(k)}, r^{(k)})$ denote the triple at iteration $k$. The refinement step is defined as:
\begin{equation}
\label{eq:refinement}
\begin{aligned}
\mathcal{E}^{(k)} &= \textsc{Diagnose}(q^{(k)}, s^{(k)}, r^{(k)}, \mathcal{K}, \mathcal{I}), \\
\Phi^{(k)} &= \textsc{Retrieve}(\mathcal{E}^{(k)}, \mathcal{K}), \\
(q^{(k+1)}, s^{(k+1)}, r^{(k+1)}) &= \textsc{Correct}(q^{(k)}, s^{(k)}, r^{(k)}, \Phi^{(k)}, \mathcal{K})
\end{aligned}
\end{equation}
where $\mathcal{E}^{(k)}$ denotes the detected error types (e.g., invalid columns, type mismatches, and incorrect joins) together with their locations in the reasoning trace $r^{(k)}$, $\Phi^{(k)}$ denotes the corrective evidence retrieved from $\mathcal{K}$, and $(q^{(k+1)}, s^{(k+1)}, r^{(k+1)})$ is the corrected triple. The process repeats until no errors are detected or a maximum iteration limit is reached.

The \textsc{Diagnose} function validates semantic constraints by cross-referencing $s^{(k)}$ against relevant layers of $\mathcal{K}$ and identifying both error types and their positions in $r^{(k)}$. Specifically, it checks whether referenced columns exist and match the expected semantics in $\mathcal{K}_4$ (Column Analysis), verifies that JOIN conditions are consistent with the foreign key structure encoded in $\mathcal{K}_6$ (Relation Analysis), and executes $s^{(k)}$ on $\mathcal{I}$ to verify executability while validating aggregation logic against type constraints in $\mathcal{K}_3$ (Field Type Analysis).

The \textsc{Retrieve} operation queries $\mathcal{K}$ using the detected errors $\mathcal{E}^{(k)}$ to extract corrective evidence $\Phi^{(k)}$. Based on this evidence, the \textsc{Correct} function synthesizes the next triple $(q^{(k+1)}, s^{(k+1)}, r^{(k+1)})$ by replacing invalid schema elements in $s^{(k)}$ with valid alternatives, reformulating $q^{(k)}$ when needed to resolve ambiguities or incorrect assumptions, and appending the detected errors together with the applied corrections to $r^{(k+1)}$ in order to preserve the refinement history.

\section{Experiments}
\label{sec:experiments}

Our empirical study is designed to answer three research questions: downstream effectiveness (RQ1), synthetic data quality (RQ2), and scaling behavior (RQ3). The remainder of this section describes the datasets, baselines, metrics, and implementation used to answer these questions.

\subsection{Datasets and Baselines}

We evaluate SemanticAgent on benchmarks spanning cross-domain parsing, domain-specific reasoning, and robustness, and compare against synthetic-data generation baselines under a controlled fine-tuning protocol.

We select three representative benchmarks—Spider, BIRD, and Spider2.0—to evaluate cross-domain parsing, knowledge-intensive reasoning, and enterprise-level SQL complexity, respectively. Spider\cite{yu2018spider} is the standard cross-domain benchmark, comprising 10,181 question–SQL pairs across 200 databases from 138 domains. BIRD\cite{li2023can} contains 12,751 question–SQL pairs across 95 databases in 37 domains; 42.8\% of questions require knowledge beyond direct schema matching, making it suitable for evaluating knowledge-intensive reasoning. Spider2.0~\cite{lei2024spider2} targets enterprise-level scenarios with large schemas and multi-step SQL workflows, representing a more challenging and realistic evaluation setting. Following prior work, we use its SQLite subset (135 samples) from the full set of 632 tasks.

We further evaluate on six benchmarks targeting domain transfer, question robustness, and domain-knowledge requirements. EHRSQL\cite{lee2022ehrsql} and ScienceBenchmark\cite{zhang2023sciencebenchmark} assess transfer to clinical and scientific domains. Spider-Syn\cite{gan2021syn} and Spider-Realistic\cite{deng2021structure} test robustness against synonym substitution and explicit column-name removal. Spider-DK~\cite{gan2021exploring} targets cases requiring domain knowledge beyond schema-level matching.

We compare SemanticAgent against three synthetic-data generation baselines under a controlled fine-tuning setting. We additionally include prompting-based systems as end-to-end reference points. CodeS\cite{li2024codes} is a template-based method that increases structural diversity across schemas. SynQL\cite{baumgartner2024synql} synthesizes SQL by decomposing and recombining abstract syntax trees. OmniSQL~\cite{li2025omnisql} combines structural transformations with LLM-based verification.

All baselines are evaluated under a matched protocol with identical data budget, downstream backbone, and training configuration to ensure a fair comparison.

\subsection{Synthetic Data Construction and Source Statistics}

\begin{table}[!t]
\centering
\caption{Database Statistics and Synthetic Corpus Sizes}
\label{tab:synth_source_stats}

\begin{tabular}{lccccc}
\toprule
Benchmark & DB & Domain & Avg. Tbl/DB & Avg. Col/DB & Synth. Pairs \\
\midrule
Spider 1.0 & 200 & 138 & 5.11 & 26.82 & 40,000 \\
Spider~2.0-SQLite & 30 & 8 & 13.8 & 7.0 & 20,000 \\
BIRD & 95 & 37 & 7.64 & 54.56 & 30,000 \\
EHRSQL & 2 & 1 & 31.55 & 92.00 & 5,000 \\
ScienceBenchmark & 3 & 3 & 16.67 & 44.47 & 5,000 \\
\bottomrule
\end{tabular}%

\end{table}

Synthetic corpora are constructed separately for each benchmark. Specifically, we generate 40,000 question–SQL pairs for Spider 1.0, 20,000 for Spider 2.0-SQLite, 30,000 for BIRD, and 5,000 each for EHRSQL and ScienceBenchmark. Spider-Syn, Spider-Realistic, and Spider-DK are reserved for evaluation only, with no synthetic data generated for them.

All synthetic data are generated exclusively from database-side information, including schema structures and sampled cell values. We do not use any gold question–SQL pairs from public benchmarks—neither as synthesis seeds nor as in-context exemplars. Table~\ref{tab:synth_source_stats} reports source statistics for each benchmark, including the numbers of databases and domains, average tables and columns per database, and the size of each synthetic corpus.

To reduce accidental contamination, we apply an n-gram overlap filter against the public evaluation split of each benchmark. Specifically, we discard any synthesized example whose question or SQL query shows high n-gram overlap with an existing evaluation instance. In total, this filter removes 764 examples (0.76\% of the initially generated corpus), leaving 99,236 examples for downstream training.

\subsection{Evaluation Metrics}
We evaluate SemanticAgent with respect to three research questions: downstream effectiveness (RQ1), synthetic data quality (RQ2), and scaling behavior (RQ3).

For RQ1, we use execution-based metrics to assess SQL correctness. Spider (dev), Spider-Syn, and Spider-Realistic are evaluated with test-suite accuracy (TS)~\cite{zhong2020semantic}, while the remaining benchmarks are evaluated with execution accuracy (EX)~\cite{yu2018spider,li2023can}. EX compares execution results on a single database, whereas TS evaluates predictions over multiple test databases.

For RQ2, we evaluate the synthesized data using four intrinsic metrics: Semantic Alignment (SA), Successful Execution Rate (SER), Query Complexity, and semantic diversity. SA measures question--SQL consistency using an LLM-based evaluator. SER measures the proportion of executable synthesized SQL queries. Query Complexity characterizes SQL structure using four AST-based difficulty levels. Semantic diversity measures distributional variation using the mean L2 distance and 1-NN distance over SQL embeddings. The prompt templates used for intrinsic evaluation are provided in Appendix~\ref{app:prompts}. For RQ3, we analyze downstream performance as a function of the synthetic data budget.

\subsection{Implementation Details}
\label{subsec:implementation_details}

For synthetic data generation and LLM-based intrinsic evaluation, we use Qwen3-235B-A22B-Instruct~\cite{yang2025qwen3}, deployed locally with vLLM~0.11.0~\cite{kwon2023efficient}. For downstream evaluation, we fully fine-tune three backbone models: Qwen2.5-Coder-7B, Qwen2.5-Coder-14B~\cite{hui2024qwen25coder}, and Qwen3-14B~\cite{yang2025qwen3}. All downstream backbones are trained on each synthetic corpus under identical settings.

We use the ms-swift framework~\cite{zhao2025swift} with DeepSpeed ZeRO Stage~3 offloading~\cite{rajbhandari2021zero} and \texttt{bfloat16} mixed precision. We optimize all downstream models with AdamW~\cite{loshchilov2019decoupled}, using $\beta_1=0.9$, $\beta_2=0.95$, and $\epsilon=10^{-8}$. The learning rate is linearly warmed up over the first $5\%$ of training steps and then annealed with a cosine schedule to $10\%$ of its peak value. Unless otherwise specified, all models are fully fine-tuned for 2 epochs with an effective global batch size of 512 and a peak learning rate of $4\times10^{-6}$. We use a maximum sequence length of 8{,}192 tokens, weight decay of 0.1, and gradient clipping of 1.0.

All local experiments are conducted on a server equipped with 8 NVIDIA A800-SXM4-80GB GPUs and 1,TB of RAM. PyTorch~2.9.1, vLLM~0.11.0 is used for high-throughput inference in data synthesis and intrinsic evaluation. The DeepSpeed~0.18.6 are used for distributed training,

\begin{table*}[t]
\centering
\caption{Downstream Performance on Text-to-SQL Benchmarks}

\label{tab:main_results}

\small
\setlength{\tabcolsep}{2.2pt}
\renewcommand{\arraystretch}{0.95}

\begin{tabular}{@{}lccccccccc@{}}
\toprule
\multirow{2}{*}{\textbf{LLM}}
& \textbf{Spider}
& \textbf{Spider}
& \textbf{BIRD}
& \textbf{Spider2.0}
& \textbf{Science}
& \textbf{EHRSQL}
& \textbf{Spider-}
& \textbf{Spider-}
& \textbf{Spider-} \\
& \textbf{(dev)}
& \textbf{(test)}
& \textbf{(dev)}
& \textbf{SQLite}
& \textbf{Bench.}
&
& \textbf{DK}
& \textbf{Syn}
& \textbf{Real.} \\
\midrule

\multicolumn{10}{c}{\textbf{LLM based Methods}} \\
\midrule
QC2.5-7B
  & 73.4 & 82.2 & 50.9 & 1.5  & 45.2 & 24.3 & 67.5 & 63.1 & 66.7 \\
QC2.5-14B
  & \textbf{78.1} & 86.6 & 61.5 & 5.9  & 52.2 & 31.6 & 73.6 & 68.2 & \textbf{76.2} \\
Q3-14B
  & 76.7 & 83.9 & 55.2 & 1.4  & 53.1 & 38.5 & 74.0 &67.1 & 71.3 \\
QC2.5-32B
  & 77.7 & \textbf{87.5} & 64.5 & 5.9  & 54.8 & 36.4 & \textbf{78.3} &  \textbf{69.9} & 72.1 \\
Q2.5-72B
  & 73.9 & 84.0 & 60.3 & 9.6  & 52.8 & 35.0 & 76.4 & 64.1 & 70.1 \\
DeepSeek-V3
  & 73.1 & 85.5 & 63.2 & \textbf{12.6} & 56.2 & 43.2 & 72.9 & 64.4 & 67.9 \\
DIN-SQL (GPT-4o)
  & 74.2 & 85.3 & 55.9 & 9.1  & 59.7 & 42.5 & 74.5 & 63.8 & 68.2 \\
CHESS (GPT-4o)
  & 76.8 & 87.2 & \textbf{66.8} & 7.7  & \textbf{61.3} & \textbf{48.2} & 77.5 & 66.9 & 72.4 \\
\midrule

\multicolumn{10}{c}{\textbf{Specialized Fine-tuned Models}} \\
\midrule
OmniSQL-7B                       & 81.2 & 87.9 & 63.9 & 10.4 & 50.2 & 34.9 & 76.1 & 69.7 & 76.2 \\
OmniSQL-14B                      & 81.4 & 88.3 & 64.2 & 10.4 & 56.9 & 39.9 & 72.9 & 69.0 & 76.4 \\
OmniSQL-32B                      & 80.9 & 87.6 & 64.5 & 11.9 & 57.2 & 42.4 & 76.1 & 69.7 & 78.1 \\
\midrule

\multicolumn{10}{c}{\textbf{Synthetic Data Synthesis Methods}} \\
\midrule
CodeS (QC2.5-7B)                 & 76.1 & 82.8 & 57.2 & 2.1  & 46.8 & 30.2 & 70.3 & 64.9 & 69.8 \\
SynQL (QC2.5-7B)                & \textbf{77.9} & 83.0 & 59.6 & 3.7  & 48.5 & 33.1 & 72.0 & 66.4 & 72.3 \\
OmniSQL (QC2.5-7B)               & 76.1 & 84.9 & 59.8 & 7.1  & 47.4 & 34.5 & 72.6 & 64.8 & \textbf{72.7} \\
\textbf{Ours (QC2.5-7B)}         & 76.3 & \textbf{86.1} & \textbf{62.4} & \textbf{9.3}  & \textbf{50.6} & \textbf{36.8} & \textbf{73.9} & \textbf{67.5} & 72.4 \\
\cmidrule{1-10}
CodeS (QC2.5-14B)                & 79.5 & 87.3 & 62.1 & 4.3  & 53.4 & 34.0 & 72.1 & 69.0 & 77.0 \\
SynQL (QC2.5-14B)               & 80.6 & 87.8 & 63.2 & 5.8  & 54.6 & 36.1 & 74.8 & 69.8 & \textbf{78.0} \\
OmniSQL (QC2.5-14B)              & \textbf{81.2} & \textbf{88.2} & 64.0 & 10.2 & 56.4 & 38.8 & 73.2 & \textbf{70.2} & 75.5 \\
\textbf{Ours (QC2.5-14B)}        & 80.1 & 86.3 & \textbf{65.4} & \textbf{10.8} & \textbf{57.8} & \textbf{40.6} & \textbf{76.5} & 70.1 & 75.2 \\
\cmidrule{1-10}

OmniSQL (Q3-14B)                 & \textbf{81.1} & 86.8 & 63.7 & 9.6  & 52.6 & 37.8 & 75.1 & 67.6 & \textbf{74.8} \\
\textbf{Ours (Q3-14B)}           & 81.0 & \textbf{88.1} & \textbf{66.0} & \textbf{13.0} & \textbf{56.5} & \textbf{40.1} & \textbf{76.8} & \textbf{69.2} & 74.0 \\
\bottomrule
\end{tabular}

\vspace{1mm}
\begin{minipage}{\textwidth}
\footnotesize
\textit{Note:} This table reports main downstream results across benchmarks spanning cross-domain, domain-specific, and robustness settings. We report results across three comparison groups: (1) Reference systems: backbone models and representative prompting-based LLMs serving as upper-bound references; (2) Published models: specialized models fine-tuned on synthetic data reported in prior work; and (3) Controlled synthesis comparison: synthetic data methods evaluated under matched data sources, backbone, and fine-tuning settings. Model abbreviations: QC2.5-xB = Qwen2.5-Coder-xB; Q2.5-72B = Qwen2.5-72B; Q3-14B = Qwen3-14B.
\end{minipage}

\vspace{-2mm}
\end{table*}

\section{Results and Discussion}
\label{sec:results}

In this section, we answer the three research questions in turn. We first examine downstream benchmark performance (RQ1), then analyze the quality of the synthesized data (RQ2), and finally study scaling behavior under increasing synthetic data budgets (RQ3). Unless otherwise stated, the best results are highlighted in \textbf{bold}.

\subsection{Downstream Benchmark Performance (RQ1)}
\label{subsec:downstream_performance}

As shown in Table~\ref{tab:main_results}, the following discussion focuses on the matched comparison among CodeS, SynQL, OmniSQL, and SemanticAgent, all fine-tuned with the same backbone and training protocol on their respective synthetic corpora. Released OmniSQL variants are included only for context, not as direct baselines.

The gains are largest on the most challenging benchmarks. On BIRD, SemanticAgent improves over OmniSQL by 2.6, 1.4, and 2.3 points under Qwen2.5-Coder-7B, Qwen2.5-Coder-14B, and Qwen3-14B, respectively. On Spider2.0-SQLite, the corresponding gains are 2.2, 0.6, and 3.4 points. These results suggest that SemanticAgent is particularly effective when schema complexity and compositional reasoning demands are high.

On Spider, the margin is smaller. SemanticAgent improves over OmniSQL in the Qwen2.5-Coder-7B and Qwen3-14B settings (86.1 vs.\ 84.9 and 88.1 vs.\ 86.8), but is slightly below OmniSQL with Qwen2.5-Coder-14B (86.3 vs.\ 88.2). This weaker separation is not surprising, given Spider’s maturity and the relatively limited headroom after synthetic fine-tuning.

The benefits of SemanticAgent also extend to specialized domains. On ScienceBenchmark, it surpasses OmniSQL by 1.4--3.9 points across the three backbone settings. On EHRSQL, the corresponding gains range from 1.8 to 2.3 points. These results suggest that the synthesized data better capture domain-specific semantics in scientific and clinical settings.

A similar pattern appears on robustness-oriented benchmarks. On Spider-Syn, SemanticAgent improves over OmniSQL by 2.7 points under Qwen2.5-Coder-7B and by 1.6 points under Qwen3-14B. On Spider-DK, it improves over OmniSQL in all three settings, by 1.3, 3.3, and 1.7 points, respectively. On Spider-Realistic, however, SemanticAgent remains competitive but does not consistently achieve the best score. This suggests that synthetic supervision improves robustness to lexical variation more clearly than robustness to natural user phrasing.

For context, SemanticAgent also remains competitive with much larger general-purpose or prompting-based systems. For example, the Qwen2.5-Coder-7B variant reaches 62.4 on BIRD, outperforming Qwen2.5-72B~\cite{yang2024qwen25} (60.3). The Qwen3-14B variant achieves 13.0 on Spider2.0, surpassing DeepSeek-V3~\cite{liu2024deepseekv3} (12.6), DIN-SQL~\cite{pourreza2023din} (9.1), and CHESS~\cite{talaei2024chess} (7.7). Because these comparisons are not fully controlled, we treat them only as reference points.

Overall, the downstream results support two conclusions. First, under matched backbones and training protocols, SemanticAgent consistently improves over prior synthetic-data construction methods. Second, the gains are largest on benchmarks that require stronger schema understanding, domain-specific reasoning, or robustness to lexical variation, indicating that higher-quality synthetic supervision is particularly valuable in semantically demanding settings.

\subsection{Synthetic Data Quality (RQ2)}

To better understand the data properties associated with SemanticAgent’s downstream gains, we analyze the synthesized data from three complementary perspectives: validity, structural complexity, and semantic diversity. Table~\ref{tab:consistency_metrics} reports two validity-related metrics, Successful Execution Rate (SER) and Semantic Alignment (SA). SER results are mixed across benchmarks. However, SemanticAgent achieves the highest SER on Spider~2.0-SQLite, the most challenging benchmark in our evaluation. This suggests stronger executability in the setting where execution is most difficult, rather than uniformly higher executability across all benchmarks.

\begin{table}[!ht]
\centering
\caption{Validity assessment of synthesized datasets}
\label{tab:consistency_metrics}

\begin{tabular}{lcccccc}
\toprule
\textbf{Method} & \multicolumn{2}{c}{\textbf{BIRD}} & \multicolumn{2}{c}{\textbf{Spider 1.0}} & \multicolumn{2}{c}{\textbf{Spider~2.0-SQLite}} \\
\cmidrule(lr){2-3} \cmidrule(lr){4-5} \cmidrule(lr){6-7}
& \textbf{SER} & \textbf{SA} & \textbf{SER} & \textbf{SA} & \textbf{SER} & \textbf{SA} \\
\midrule
CodeS & 91.37 & \underline{95.12} & 92.73 & \underline{92.47} & 21.06 & 89.94 \\
SynQL & 90.12 & 92.16 & \textbf{94.18} & 90.06 & 25.78 & \textbf{94.18} \\
OmniSQL & \textbf{94.83} & \textbf{97.41} & \underline{93.28} & 91.65 & \underline{35.31} & \underline{90.39} \\
SemanticAgent & \underline{92.14} & 91.31 & 93.18 & \textbf{94.37} & \textbf{63.13} & 72.17 \\
\bottomrule
\end{tabular}%

\begin{minipage}{0.7\textwidth}
\footnotesize
\vspace{1mm}
\textit{Note:} SER and SA denote Successful Execution Rate and Semantic Alignment, respectively. Higher values indicate better validity for both metrics.
\end{minipage}

\end{table}

\begin{figure*}[!t]
    \centering
    \subfloat[Complexity distributions of synthetic SQL queries across synthesis methods.]{
        \includegraphics[width=0.45\textwidth]{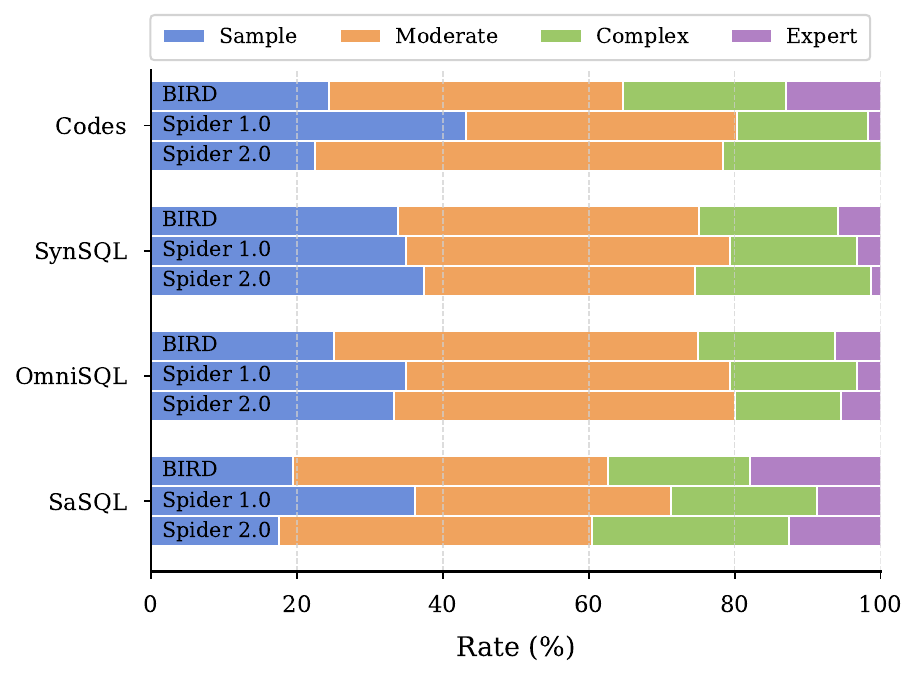}
        \label{fig:data_quality}
    }
    \hfill
    \subfloat[Performance scaling of Qwen3-14B with increasing amounts of SemanticAgent-generated synthetic data.]{
        \includegraphics[width=0.45\textwidth]{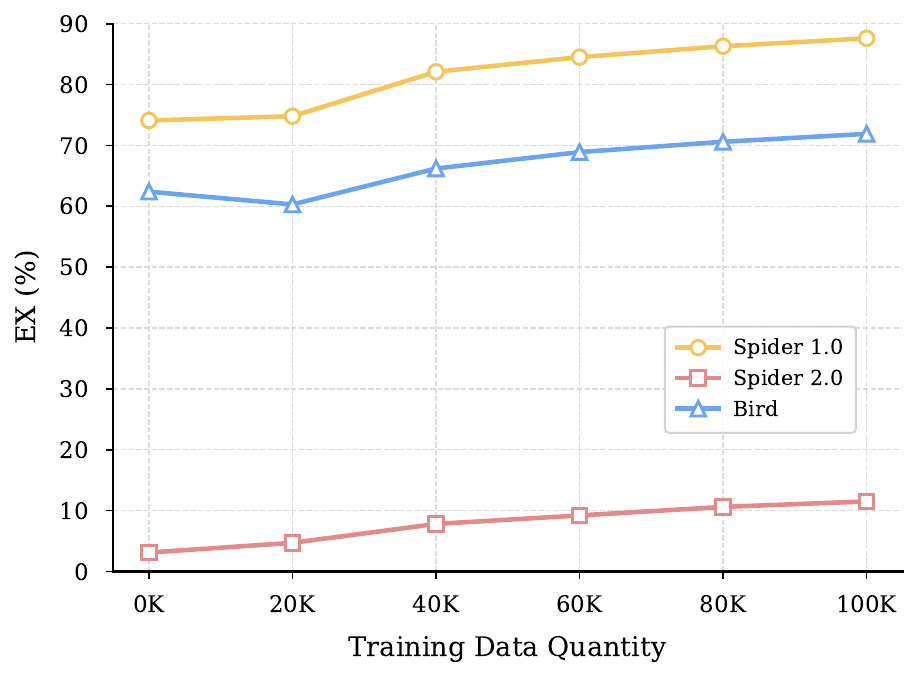}
        \label{fig:scaling_law}
    }
    \caption{Analysis of synthesized data from the perspectives of structural quality and data efficiency. 
    (a) SemanticAgent generates a larger proportion of moderately complex and complex SQL queries, indicating broader structural coverage. 
    (b) Performance scaling of Qwen3-14B with increasing amounts of SemanticAgent-generated synthetic data. 
    As the amount of synthetic data increases, downstream performance improves consistently, demonstrating strong data efficiency and favorable scaling behavior.}
    \label{fig:analysis_quality_scaling}
\end{figure*}

Although SemanticAgent obtains lower SA on Spider~2.0 than some baselines, this result should be interpreted together with SER. SA is computed only over executable SQL queries, rather than over the full set of generated outputs. Because baseline methods achieve substantially lower SER on Spider~2.0, their SA is evaluated on smaller executable subsets. As shown in Figure~\ref{fig:analysis_quality_scaling}(a), these subsets may be dominated by simpler queries. Such queries are more likely to align structurally with the reference SQL and may therefore yield inflated SA scores. Under this evaluation protocol, a higher SA does not necessarily imply higher overall data quality.

Figure~\ref{fig:analysis_quality_scaling}(a) shows that SemanticAgent generates a larger proportion of moderately complex and complex SQL queries. This indicates broader structural coverage. Table~\ref{tab:diversity_metrics} reports semantic diversity in the embedding space. SemanticAgent achieves the highest Mean L2 and 1-NN scores on all three benchmarks. This suggests lower redundancy and broader coverage of the synthesized query space.

Taken together, these results indicate that SemanticAgent improves synthesized data quality along multiple dimensions. In addition to substantially stronger executability on the most challenging benchmark, it generates structurally broader and more complex SQL queries and yields higher semantic diversity across datasets. This combination may help explain the larger downstream gains observed on more challenging benchmarks.

\begin{table}[ht]
\centering
\caption{Semantic diversity of synthesized datasets}
\label{tab:diversity_metrics}

\begin{tabular}{@{}lcccccc@{}}
\toprule
\textbf{Method} & \multicolumn{2}{c}{\textbf{BIRD}} & \multicolumn{2}{c}{\textbf{Spider 1.0}} & \multicolumn{2}{c}{\textbf{Spider~2.0-SQLite}} \\
\cmidrule(lr){2-3} \cmidrule(lr){4-5} \cmidrule(lr){6-7}
& \textbf{Mean L2} & \textbf{1-NN} & \textbf{Mean L2} & \textbf{1-NN} & \textbf{Mean L2} & \textbf{1-NN} \\
\midrule
CodeS & 0.6213 & 0.2441 & 0.6152 & 0.2394 & 0.4086 & 0.2359 \\
SynQL & 0.6421 & 0.2553 & 0.6354 & 0.2519 & 0.5289 & 0.2175 \\
OmniSQL & \underline{0.6623} & \underline{0.2681} & \underline{0.6568} & \underline{0.2421} & \underline{0.5497} & \underline{0.2210} \\
SemanticAgent & \textbf{0.6819} & \textbf{0.2745} & \textbf{0.6737} & \textbf{0.2645} & \textbf{0.6634} & \textbf{0.2792} \\
\bottomrule
\end{tabular}%

\begin{minipage}{0.5\textwidth}
\footnotesize
\vspace{1mm}
\textit{Note:} Higher values indicate greater diversity.
\end{minipage}
\end{table}

\subsection{Scaling Behavior (RQ3)}

We further examine how downstream performance scales with the synthetic data budget. Figure~\ref{fig:analysis_quality_scaling}(b) shows that SemanticAgent consistently outperforms baseline synthesis methods as the data budget increases. Its advantage emerges early in the low-resource regime and remains stable as the data volume grows. This suggests better data efficiency and more favorable scaling behavior in this setting.

The early improvement in the low-data regime is particularly notable. It suggests that SemanticAgent provides more informative training instances. As a result, the model can acquire useful SQL patterns with fewer synthesized examples. This interpretation is consistent with the data-quality analysis above, which shows higher semantic diversity and broader complexity coverage.

As expected, all methods exhibit diminishing returns as the training set becomes larger. Nevertheless, SemanticAgent maintains the strongest performance curve throughout the scaling analysis. This suggests that, in this setting, improving the intrinsic quality of synthetic data may be more beneficial than increasing its quantity alone.

\section{Conclusion}

In this paper, we identify a key limitation of existing text-to-SQL synthesis pipelines: execution success does not guarantee semantic correctness. Synthetic samples may execute successfully while still violating schema semantics or domain-specific constraints, which can reduce synthetic data quality and limit downstream performance. To address this issue, we propose \textbf{SemanticAgent}, a knowledge-guided synthesis framework built on a structured knowledge base derived from database schemas and data. This knowledge base supports both question--SQL generation and sample verification. By validating generated pairs against schema semantics and domain-specific constraints, SemanticAgent reduces semantically inconsistent yet executable samples and improves the quality of synthesized training data.

Under matched backbones and training protocols, models fine-tuned on SemanticAgent-generated data consistently outperform those trained on prior synthetic corpora. The gains are largest on BIRD and Spider~2.0-SQLite, where stronger schema grounding and domain-specific reasoning are required. These results suggest that knowledge-guided verification is especially beneficial in semantically demanding text-to-SQL settings.

Several limitations remain. First, offline data synthesis is computationally expensive. Constructing the full training corpus requires approximately 2{,}880 GPU-hours, with verification and refinement accounting for 74.3\% of the total cost. Second, the current pipeline relies on a single teacher model, so bias in that model or contamination from its pretraining may propagate to the synthesized data. Third, our evaluation is limited to public academic benchmarks and may not fully capture the schema noise, SQL dialect variation, and deployment constraints of proprietary enterprise environments. Finally, we primarily use execution accuracy as the evaluation metric, which may still reward semantically incorrect SQL when it happens to produce the correct output.

Future work should focus on making verification and refinement more efficient, reducing dependence on a single teacher model, and evaluating knowledge-guided synthesis under noisier schemas and more realistic deployment settings. More broadly, our findings suggest that executability alone is insufficient for reliable synthetic supervision in text-to-SQL, and that explicit semantic verification is a promising direction for improving downstream performance after fine-tuning.

\section*{Acknowledgments}

\section*{Declaration of competing interest}
The authors declare that they have no known competing financial interests or personal relationships that could have appeared to influence the work reported in this paper.

\section*{Declaration of generative AI and AI-assisted technologies in the writing process}
During the preparation of this work the authors used ChatGPT in order to improve writing. After using this tool/service, the authors reviewed and edited the content as needed and take full responsibility for the content of the publication.

\bibliographystyle{unsrt}
\bibliography{references}

\clearpage
\appendix

\subsection{ER Analysis}
\label{app:er-analysis}

% To fit Elsevier 1p single-column layout, the ER content is split into three compact tables with fixed column widths (p{...}), tight spacing, and small font for readability.

\begin{table*}[!ht]
\centering
\small
\setlength{\tabcolsep}{4pt}
\renewcommand{\arraystretch}{1.08}
\caption{Entity identifiers, categorical descriptors, and quantitative indicators for text-to-SQL synthesis.}
\label{tab:er-indicators}
\begin{tabular}{@{}llp{0.62\linewidth}@{}}
\toprule
\textbf{Entity} & \textbf{Type} & \textbf{Description} \\
\midrule
School & CD & School category; district; county; charter status \\
School & CD & Geographic coordinates (latitude, longitude) \\
School & CD & Magnet program; operational status (active) \\
School & QI & Quantitative indicators tracked under AP/SE entities \\
\midrule
SE & CD & IRC category code (e.g., \texttt{1}) \\
SE & CD & NSLP provision status \\
SE & CD & CALPADS certification status \\
SE & QI & FRPM eligibility rate (ages 5--17) \\
\midrule
AP & CD & Subject areas (reading, mathematics, writing) \\
AP & QI & Mean reading score; mean mathematics score \\
AP & QI & Mean writing score; count of scores (\(\geq 1500\)) \\
AP & QI & Participation rate (\texttt{NumTestTakers}/\texttt{Grade12Enrollment}) \\
\midrule
ADM & CD & Opening date; closing date; charter ID number \\
ADM & CD & Virtual campus indicator; funding type \\
ADM & QI & District of Choice (DOC), e.g., \texttt{code 00} \\
\bottomrule
\end{tabular}
\end{table*}

\begin{table*}[!ht]
\centering
\footnotesize
\setlength{\tabcolsep}{3pt}

\begin{minipage}[t]{0.30\textwidth}
    \centering
    \caption{Entity functional roles.}
    \label{tab:er-core}
    \begin{tabular}{@{}ll@{}}
        \toprule
        \textbf{Entity} & \textbf{Role} \\
        \midrule
        SCH & Primary entity \\
        SE  & Domain attribute \\
        AP  & Domain attribute \\
        ADM & Metadata entity \\
        \bottomrule
    \end{tabular}
\end{minipage}
\hfill
\begin{minipage}[t]{0.35\textwidth}
    \centering
    \caption{Source tables by entity.}
    \label{tab:er-sources}
    \begin{tabular}{@{}lp{0.72\linewidth}@{}}
        \toprule
        \textbf{Entity} & \textbf{Source Tables} \\
        \midrule
        SCH & \texttt{schools}; \texttt{satscores}; \texttt{frpm} \\
        SE  & \texttt{frpm}; \texttt{cdscode}; \texttt{AY} \\
        AP  & \texttt{satscores}; \texttt{cdscode} \\
        ADM & \texttt{schools}; \texttt{cdscode} \\
        \bottomrule
    \end{tabular}
\end{minipage}

\begin{minipage}{0.9\textwidth}
\footnotesize
\vspace{1mm}
\textit{Note:} 
\noindent Entity abbreviations used throughout this analysis: School (\textbf{SCH}), Socioeconomic eligibility (\textbf{SE}), Academic performance (\textbf{AP}), \texttt{Academic year} (\texttt{AY}), Administrative metadata (\textbf{ADM}), Categorical descriptors (\textbf{CD}) and Quantitative indicators (\textbf{QI}).
\end{minipage}
\end{table*}

\clearpage
\subsection{Synthetic data format with structured reasoning}
\label{app:training_format}

\lstdefinestyle{jsonStyle}{
    basicstyle=\scriptsize\ttfamily,
    numbers=none,
    breaklines=true,
    breakatwhitespace=true,
    tabsize=2,
    showstringspaces=false,
    columns=flexible,
    keepspaces=true,
    frame=none
}

\begin{figure*}[!ht]
\centering
\begin{tcolorbox}[
  width=.7\linewidth,
  enhanced,
  boxrule=0.5pt,
  colback=AliceBlue,
  colframe=Navy,
  colbacktitle=Navy,
  fonttitle=\bfseries\color{white}, 
  title=Synthetic data format with structured reasoning,
  arc=2mm,
  attach boxed title to top left={yshift=-2.5mm, xshift=3mm},
  boxed title style={arc=1mm},
  boxsep=5pt,
  left=8pt,
  right=8pt,
  top=6mm,
]
\begin{lstlisting}[style=jsonStyle]
{
  "question": "Which school has the highest average SAT math score?",
  "think": {
    "focus": "Identify single school with MAX(AvgScrMath)",
    "metadata": {
      "main_scenario": "ranking_query",
      "sub_scenario": "max_min_query",
      "complexity_level": 1,
      "use_case": "ranking_analysis"
    },
    "table_selection": {
      "tables_used": ["satscores"],
      "reasoning": "Contains SAT columns for extreme value. Single-table compliant with Level 1"
    },
    "column_selection": {
      "columns_used": [
        {
          "name": "sname",
          "type": "TEXT",
          "operation": "SELECT",
          "purpose": "Identify school with extreme score"
        },
        {
          "name": "AvgScrMath",
          "type": "INTEGER",
          "operation": "ORDER BY DESC",
          "purpose": "Find maximum via sorting"
        }
      ]
    },
    "sql_strategy": {
      "operations": ["SELECT", "ORDER BY DESC", "LIMIT"],
      "approach": "ORDER BY + LIMIT for extreme value, no aggregation",
      "no_need": ["JOIN", "GROUP BY", "MAX()"]
    },
    "expected_output": "Single row: school name with highest AvgScrMath"
  },
  "answer": "SELECT sname FROM satscores ORDER BY AvgScrMath DESC LIMIT 1;"
}
\end{lstlisting}
\end{tcolorbox}
\caption{Synthetic data format with structured reasoning.}
\label{fig:training_data_format}
\end{figure*}

\clearpage

\subsection{Prompts}
\label{app:prompts}

\begin{figure*}[!ht]
\centering
\begin{tcolorbox}[
  enhanced,
  boxrule=0.5pt,
  colback=AliceBlue,
  colframe=Navy,
  colbacktitle=Navy,
  fonttitle=\bfseries\color{white}, 
  title=Prompt for SQL Complexity Classification,
  arc=2mm,
  attach boxed title to top left={yshift=-2.5mm, xshift=3mm},
  boxed title style={arc=1mm},
  boxsep=5pt,
  left=8pt,
  right=8pt,
  top=6mm,
]
\scriptsize\ttfamily  
You are an SQL complexity evaluator. Classify the given SQL query into one of four levels based on the following criteria. Output only the level number and reasoning.

\vspace{1.5ex}
\textbf{Classification Criteria:}

\vspace{0.5ex}
\textbf{Level 1: Simple}\\
Single table, basic filtering and sorting (WHERE, ORDER BY, LIMIT, DISTINCT)

\vspace{0.5ex}
\textbf{Level 2: Moderate}\\
Multi-table joins (max 3 tables, 2 JOINs) with aggregation (GROUP BY, HAVING, COUNT/SUM/AVG)

\vspace{0.5ex}
\textbf{Level 3: Complex}\\
Advanced logic: subqueries (IN, EXISTS, correlated), CASE WHEN, UNION, or 4+ tables

\vspace{0.5ex}
\textbf{Level 4: Expert}\\
Must use: Window functions (ROW\_NUMBER, RANK, LAG, LEAD), CTEs (WITH), or recursive queries

\vspace{1.5ex}
\textbf{Decision Rule:}\\
Check Level 4 features first $\rightarrow$ Level 3 $\rightarrow$ Level 2 $\rightarrow$ else Level 1

\vspace{1.5ex}
\textbf{Output JSON:}\\
\{``level'': ``number'', ``reasoning'': ``brief explanation''\}

\vspace{1.5ex}
\textbf{SQL Query:}\\
\{sql\_query\}
\end{tcolorbox}
\caption{The prompt used for SQL complexity classification. The model classifies queries into four levels based on their features and constraints.}
\label{fig:prompt_sql_complexity}
\end{figure*}

\begin{figure*}[!ht]
\centering
\begin{tcolorbox}[
  enhanced,
  boxrule=0.5pt,
  colback=AliceBlue,
  colframe=Navy,
  colbacktitle=Navy,
  fonttitle=\bfseries\color{white}, 
  title=Prompt for Question-SQL Semantic Consistency Verification,
  arc=2mm,
  attach boxed title to top left={yshift=-2.5mm, xshift=3mm},
  boxed title style={arc=1mm},
  boxsep=5pt,
  left=8pt,
  right=8pt,
  top=6mm,
]
\scriptsize\ttfamily
You are a semantic consistency evaluator. Determine whether the given SQL query correctly answers the natural language question based on the provided database schema. Output 1 if consistent, 0 if inconsistent.\\[2ex]

\textbf{Evaluation Criteria:}\\[1ex]
\textbf{Consistent (1):} The SQL query accurately retrieves the information requested in the question, using correct tables, columns, joins, filters, and aggregations according to the schema.\\[1ex]

\textbf{Inconsistent (0):} The SQL query has one or more issues:\\
- Targets wrong tables or columns\\
- Uses incorrect join conditions or missing necessary joins\\
- Applies wrong filters, aggregations, or ordering\\
- Returns irrelevant or incomplete data for the question\\[2ex]

\textbf{Verification Steps:}\\
1. Parse the question to identify required information\\
2. Check if SQL uses correct tables/columns from schema\\
3. Verify joins, filters, and aggregations match the question intent\\
4. Confirm the output answers the question completely\\[2ex]

\textbf{Output:} \{``label'': 0 or 1, ``reasoning'': brief explanation\}\\[2ex]

\textbf{Database Schema:}\\
\{schema\}\\[2ex]

\textbf{Natural Language Question:}\\
\{question\}\\[2ex]

\textbf{SQL Query:}\\
\{sql\_query\}
\end{tcolorbox}
\caption{The prompt used for Question-SQL semantic consistency verification. The model evaluates whether the SQL query correctly answers the question given the database schema.}
\label{fig:prompt_consistency}
\end{figure*}

\clearpage

 

% \end{IEEEbiography}



\vfill

\end{document}